\documentclass{article}
\usepackage[utf8]{inputenc}
\usepackage[english]{babel}

\usepackage[noadjust]{cite}
\usepackage{amsmath,amssymb,amsfonts}
\usepackage{algorithmic}
\usepackage{graphicx}
\usepackage{textcomp}
\usepackage{xcolor}
\def\BibTeX{{\rm B\kern-.05em{\sc i\kern-.025em b}\kern-.08em
    T\kern-.1667em\lower.7ex\hbox{E}\kern-.125emX}}

\usepackage{hyperref}
\usepackage{tabularx}

\usepackage{blkarray}

\title{GERNERMED - An Open German Medical NER Model}

\author{
  Johann Frei
  \and
  Frank Kramer
  \and
  \texttt{firstname.lastname@informatik.uni-augsburg.de}
}
\date{September 2021}

\begin{document}

\maketitle

\begin{abstract}

The current state of adoption of well-structured electronic health records and integration of digital methods for storing medical patient data in structured formats can often be considered as inferior compared to the use of traditional, unstructured text based patient data documentation. Data mining in the field of medical data analysis often needs to rely solely on processing of unstructured data to retrieve relevant data. In natural language processing (NLP), statistical models have been shown successful in various tasks like part-of-speech tagging, relation extraction (RE) and named entity recognition (NER).

In this work, we present \textit{GERNERMED}, the first open, neural NLP model for NER tasks dedicated to detect medical entity types in German text data. Here, we avoid the conflicting goals of protection of sensitive patient data from training data extraction and the publication of the statistical model weights by training our model on a custom dataset that was translated from publicly available datasets in foreign language by a pretrained neural machine translation model.

The sample code and the statistical model is available at:\\\url{https://github.com/frankkramer-lab/GERNERMED}

\end{abstract}

\section{Introduction}
Despite continuous efforts to transform the storage and processing of medical data in healthcare systems into a framework of machine-readable highly structured data, implementation designs that aim to fulfill such requirements are only slowly gaining traction in the clinical healthcare environment. In addition to common technical challenges, physicians tend to bypass or completely avoid inconvenient data input interfaces, which enforce structured data formats, by encoding relevant information as free-form unstructured text.\cite{clinicaltextdatareview2020,modernclinicaltextmining}

Electronic data capturing systems are developed in order to improve the situation of structured data capturing, yet their primary focus lies on clinical studies and the involvement of these systems needs to be designed in early stages and requires active software management and maintenance. Such electronic data capturing solutions are commonly considered in the context of clinical research but are largely omitted in non research-centric healthcare services and paper-based solutions are preferred.\cite{digitalizationgermanyjmir2020,aerzteblattDigitalization2017,clinicaltextdatareview2020,modernclinicaltextmining} 

In the light of the rise of data mining and big data analysis, the emerging importance of large scale data acquisition and collection for finding and understanding novel relationships of disease, disease-indicating biomarkers, drug effects and other input variables, induces additional pressure on finding new possible data sources. While new datasets can be designed and created for specific use cases, the amount of obtained data might be very limited and not sufficient for modern data-driven methods. Furthermore, such data collection efforts can turn out as rather inefficient in terms of time and work involved in creating new datasets with respect to the number of acquired data samples.

In contrast, unstructured data of sources from legacy systems and non research-centric healthcare, referred to as \textit{second use}, offer a potential alternative. However, techniques for information extraction and retrieval, mainly from the NLP domain, needs to be applied to transform raw data into structured information.

While the availability of existing NLP models in English, and other non NLP-based techniques, for medical use cases is focus of active research, the situation of medical NLP models for non-English languages is less satisfying. Since the performance of an NLP model often depends on its dedicated target language, most models cannot be shared and reused easily on different languages, but requires re-training on new data from the desired target language.

In particular, for the case of detection of entities like prescribed drugs and level or frequency of dosage from German medical documents like doctoral letters, no open and publicly available model has been published to the best of our knowledge. We attribute two main contributing factors specifically to this fact:
\begin{itemize}
    \item \textbf{Lack of public German datasets:} Most open public datasets are designed for English data only. Until recently, no German dataset has been published. Specifically in the context of clinical data, legal restrictions and privacy policies prevent collection and publication of German datasets. Data-driven NLP research for medical applications utilize largely internal data for training and evaluation. In addition to the dataset itself, in order to model relevant text features with supervised learning, high quality annotations of the dataset are essential for robust model performance.
    \item \textbf{Protection of sensitive data and privacy concerns:} While few works have been published that present data-driven models for German texts, the weights of these models have not been published. Since respective training data has been used in non-anonymized or pseudonymized fashion, the publication of the model weights inherently comes at the risk of possible data leakage issues through training data extraction from the model, potentially exposing sensitive information like patient names or id numbers.
\end{itemize}

In this paper, we aim to tackle the issue of absent of anonymous training data as well as publicly available medical German NLP models. Our main contributions are as follows:

\begin{itemize}
    \item \textbf{Automated retrieval of German dataset:} We create a custom dataset for our target language, based on a public English dataset. In addition, we apply a strategy to preserve relevant annotation information across languages.
    \item \textbf{Training of medical German NLP model component:} We trained and built a named entity recognition component on the custom dataset. The model pipeline supports multiple types of medical entities.
    \item \textbf{Evaluation and publication of the NLP pipeline:} The NER model was evaluated as part of an NLP pipeline. The trained model is publicly available for further use by third parties.
\end{itemize}

\section{Materials and Methods}
\subsection{Related Work}

In recent years substantial progress has been in the area of NLP which can mostly be attributed to the joint use of large amounts of data and its processing through large language models like BERT\cite{BERT2019} and its (bio)medical-specific models\cite{BioMedTransferBLUEBench2019,MedBERT2021,BioBERT2019,ClinicalBERT2019,SciBERT2019,ehrBERT2019} as a straightforward way to encode representations of semantic information for further processing in downstream tasks like text classification or text segmentation. These works mostly focus on English language due to available language corpora like scientific texts from PubMed or specifically designed corpora such as n2c2\cite{n2c22018} (with annotations), MIMIC-III\cite{mimiciii2016}. For German, only GGPONC\cite{GGPONC2020} has been published during our work on our project as a dataset that carries annotation information, yet other German datasets\cite{CLEF2020,Jena2018} do not. Moreover, the \textit{Technical-Laymen}\cite{Seiffe2020} corpus provides an annotated corpus, yet it is based on crawled texts from non-professional online forums.
Various other German medical text corpora exist\cite{FRAMED2004,Fette2012,Bretschneider2013,toepfer2015,lohr1992,kreuzthaler2015,roller2016,cotik2016,krebs2017,hahn2018,minarro2019,konig2019} as basis for certain NLP and information extraction use-cases, but are inaccessible for public distribution.

In the field of NLP systems for German medical texts, \textit{medSynDiKATe}\cite{Medsyndikate2002,HAHN1999} approaches information extraction on pathological finding reports by parsing and mapping text elements to (semi)automatically built knowledge representation structures.
Processing of pathological findings in German has also been applied for the tasks of sentence classification\cite{Bretschneider2013}.
In the context of patient records, a hybrid RE and NER parsing approach using the \textit{SProUT}\cite{SProUT2004} parser has been proposed\cite{krieger2014}, however the entity tags lack medical relevance. Similar general NER for non-medical entity tags has been applied in order to enable de-identification of clinical records\cite{richter2018ner} using statistical and regex-based models through the StanfordNLP parser\cite{CoreNLP2014}.

Neural methods have been shown to perform well on certain NLP tasks. In particular, convolutional (CNN) approaches for RE \cite{nguyen2015relation,sahu2016relation,zeng2014relation} have become popular in recent years. For German texts, the performance of various methods have been investigated for medical NER tasks\cite{roller2017ner}, such as CNN, LSTM or SVM-based models. In this context, the text processing platform \textit{mEx}\cite{mEx2018} uses CNN-based methods for solving medical NER in German texts. Similar to our work, \textit{mEx} is build on \textit{SpaCy}\cite{spacy}, but provides custom models for other NLP tasks such as RE. However, the platform has been partially trained on internal clinical data and thus, its statistical models have not been openly published and may only be used under certain legal restrictions.

\subsection{Custom Dataset Creation}
We rely on the publicly available training data from \textit{n2c2 NLP 2018 Track 2}\cite{n2c22018} dataset \textit{(ADE and Medication Extraction Challenge)} as our initial source dataset. The data is composed of 303 annotated text documents that have been postprocessed by the editor for anonymization purposes in order to explicitly mask sensitive privacy-concerning information.

In order to transform the data into a semantically plausible text, we identify the type and text span of text masks such that we are able replace the text masks by sampling type-compatible data randomly from a set of sample entries. During the sampling stage, depending on the type of the mask, text samples for entities like dates, names, years or phone numbers are generated and inserted into the text. Since every replacement step might affects the location of the text annotation labels as provided by the character-wise start and stop indices, these label annotation indices must be updated accordingly. For a further preprocessing, we split up the text into single sentences such that we can omit all sentences with no associated annotation labels.

For automated translation, we make use of the open source \textit{fairseq}\cite{fairseq2019} (0.10.2) model architecture. \textit{fairseq} is an implementation of a neural machine translation model, which supports automatic translation of sequential text data using pretrained models. For our purposes, we ran the \textit{transformer.wmt19.en-de} pretrained model to translate our set of English sentences into German.

The reconstructive mapping of the annotation labels from the English source text to the German target text is tackled by \textit{FastAlign}\cite{fastalign}. \textit{FastAlign} is an unsupervised method for aligning words from two sentences of source and target language. We project the annotation labels onto the translated German sentences using the word-level mapping between the corresponding English and German sentence in order to obtain new annotation label indices in the German sentence.

The word alignment mapping tends to induce errors in situations of sentences with irregular structure such as tabular or itemized text sections. We mitigate the issue and potential subsequent error propagation by inspecting the structure of the word mapping matrix $A$.
\[
A_{regular} =
\begin{blockarray}{ccccccc}
    \text{} & \text{The} & \text{cat} & \text{sat} & \text{on} & \text{the} & \text{mat.} \\
    \begin{block}{r(cccccc)}
      \text{Die}       & 1 & 0 & 0 & 0 & 0 & 0 \\
      \text{Katze}     & 0 & 1 & 0 & 0 & 0 & 0 \\
      \text{sa\ss}     & 0 & 0 & 1 & 0 & 0 & 0 \\
      \text{auf}       & 0 & 0 & 0 & 1 & 0 & 0 \\
      \text{der}       & 0 & 0 & 0 & 0 & 1 & 0 \\
      \text{Matte.}    & 0 & 0 & 0 & 0 & 0 & 1 \\
    \end{block}
\end{blockarray}
\]

In situations where \textit{FastAlign} fails to establish a meaningful mapping between source and target sentence, it can be observed that the resulting mapping table collapses to a highly non-diagonal matrix structure as illustrated by the following example:
\[
A_{irregular} =
\begin{blockarray}{ccccccc}
    \text{} & \text{The} & \text{cat} & \text{sat} & \text{on} & \text{the} & \text{mat.} \\
    \begin{block}{r(cccccc)}
      \text{Die}       & 0 & 0 & 0 & 0 & 0 & 0 \\
      \text{Katze}     & 1 & 1 & 0 & 0 & 0 & 0 \\
      \text{sa\ss}     & 0 & 0 & 1 & 0 & 0 & 0 \\
      \text{auf}       & 0 & 0 & 0 & 0 & 0 & 0 \\
      \text{der}       & 0 & 0 & 0 & 0 & 0 & 0 \\
      \text{Matte.}    & 0 & 0 & 0 & 1 & 1 & 1 \\
    \end{block}
\end{blockarray}
\]

Severely ill-aligned word mapping matrices can be detected and removed from the final set of sentences by applying the simple filter decision rule

\begin{equation}
    \dfrac{
    \sum^{w_{de}}_{i=1} \sum^{w_{en}}_{j=1} A_{ij}
    \dfrac{\left| w_{en} - i*w_{en} + i - w_{de} + j*w_{de} - j \right|}
    {\sqrt{ (w_{en}-1)^2 + (w_{de}-1)^2 }}
    }{
    \max(w_{en}, w_{de})
    } > t
\end{equation}

where the average distance between a non-zero entry and the diagonal line from $A_{1,1}$ to $A_{w_{de},w_{en}}$ is evaluated, given $w_{en}$ as the number of words in the English sentence and $w_{de}$ as the number of words in the German sentence. If the value exceeds the threshold $t$, the sentence pair is disregarded for the final set of sentences.

The word mapping matrices describe a non-symmetric cross-correspondence between two language-dependent tokensets, which enables the projection of tokens within the English annotation span onto the semantically corresponding tokens in the German translation text. Therefore, the annotation label indices for the English text can be resolved to the actual indices for the translated German text at character level. Since the entity classes remain unchanged, the following annotation label types can be obtained: \textit{Drug}, \textit{Route}, \textit{Reason}, \textit{Strength}, \textit{Frequency}, \textit{Duration}, \textit{Form}, \textit{Dosage} and \textit{ADE}.

\subsection{NLP Model for NER Training}

For the buildup of our NER model as part of an NLP pipeline, we use \textit{SpaCy} as an NLP framework for training and inference.


The SpaCy NER model follows an transducer-based parsing approach instead of a state-agnostic token tagging approach.

\textbf{Embedding}: The word tokens are embedded by Bloom embeddings where different linguistic features are concatenated into a single vector and passed through $n_{embed}$ separate dense layers, followed by a final maxpooling and layer norm step. This step enables the model to learn meaningful linear combinations of single input feature embeddings while reducing the number of dimensions.

\textbf{Context-aware Token Encoding}: In order to extract context-aware features that are able to capture larger token window sizes, the final token embedding is passed through an multi-layered convolutional network. Each convolution step consists of the convolution itself as well as the following maxpooling operation to keep the dimensions constrained. For each convolution step, a residual (skip) connection is added to allow the model to pass intermediate data representations from previous layers to subsequent layers.

\textbf{NER Parsing}: For each encoded token, a corresponding feature-token vector is precomputed in advance by a dense layer. For parsing, the document is processed token-wise in a stateful manner. For NER, the state at a given position consists of the current token, the first token of the last entity and the previous token by index. Given the state, the feature-position vectors are retrieved by indexing the values from the precomputed data and sumed up. A dense layer is applied to predict the next action. Depending on the action, the current token is annotated and the next state is generated until the entire document has been parsed.

\section{Results}

\subsection{Custom Dataset Creation}

As initial preprocessing step, we need to replace the anonymization masks by meaningful regular text data to reconstruct the natural appearance of the text and alleviate a potential dataset bias that leads to gaps between the dataset and real world data. For numerical data, we can retrieve mask replacements by random sampling. Similar to numerical data, dates and years are sampled and formatted to common date formats. For semantically relevant data types, we use the Python package \textit{Faker}. The package maintains lists of plausible data of various types such as first names, last names, addresses or phone numbers. We make use of these data entries for certain typed anonymization masks.
In order to obtain our custom dataset, we split the texts from the original dataset into single sentences using the sentence splitting algorithm from SpaCy. The English sentences were translated into German by the \textit{fairseq} library with beam search (b=5). The sentence-wise word alignments were obtained by \textit{FastAlign} and cleaned up by our filter decision rule (t=1.8).

The labels \textit{Reason} and \textit{ADE} were removed from the dataset due to the fact that their definitions are rather ambiguous in general contexts beyond the scope of the initial source dataset.

Our final custom dataset consists of 8599 sentence pairs, annotated with 30233 annotations of 9 different class labels. The different class labels and their corresponding frequency in absolute numbers are shown in table \ref{tab:tags}. The German sentences consist of 172695 tokens in total.

\begin{table}[!ht]
    \centering
    \begin{tabular}{l|r}
        \hline
        \textbf{NER Tag} & \textbf{Count} \\
        \hline
        Drug         & 8305          \\
        Route        & 4071           \\
        Strength     & 4549           \\
        Frequency    & 4238           \\
        Duration     & 409            \\
        Form         & 5242           \\
        Dosage       & 3419           \\
        \hline
    \end{tabular}
    \caption{The distribution of annotations in the custom dataset in absolute numbers. The dataset consists of 8599 sentence samples (172695 tokens). A single tag sample may span multiple tokens.}
    \label{tab:tags}
\end{table}

\subsection{NLP Model for NER Training}

For training, we utilize our custom German dataset as our training data and split the dataset into training set (80\%, 6879 sentence samples), validation set and test set (both 10\%, 860 samples). The training setup follows the default NER setup of SpaCy, the Adam optimizer with a learning rate of 0.001 with decay ($\beta_{1}$ = 0.9, $\beta_{2}$ = 0.999) is used.
The training took 10 minutes on an Intel i7-8665U CPU.

\begin{figure}[!ht]
    \centering
    \includegraphics[width=\textwidth]{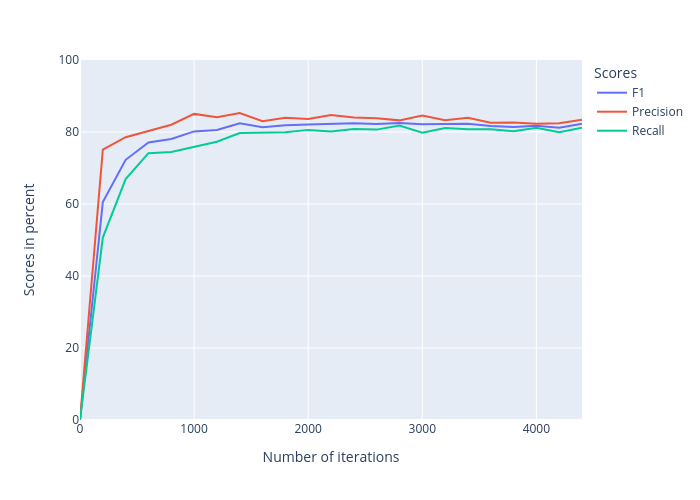}
    \caption{Training scores on validation set: Evaluation scores are computed at every $200^{th}$ iteration.}
    \label{fig:training_performance}
\end{figure}

The model performance during training is shown in figure \ref{fig:training_performance}. The corresponding performance scores are evaluated on the validation set.

We select the final model based on the highest F1-score on the validation set. The performance of the selected model is evaluated on the test set per NER tag as well as in total. The results are shown in table \ref{tab:tags_scores}.

\begin{table}[!ht]
    \centering
    \begin{tabular}{l|rrr}
        \hline
        \textbf{NER Tag} & \textbf{Precision} & \textbf{Recall} & \textbf{F1-Score} \\
        \hline
        Drug        & 67.33   & 66.17   & 66.74 \\
        Strength    & 92.34   & 90.99   & 91.66 \\
        Route       & 89.93   & 90.14   & 90.04 \\
        Form        & 91.94   & 89.24   & 90.57 \\
        Dosage      & 87.83   & 87.57   & 87.70 \\
        Frequency   & 79.14   & 76.92   & 78.01 \\
        Duration    & 67.86   & 52.78   & 59.37 \\
        \hline\hline
        total       & 82.31   & 80.79   & 81.54 \\
        \hline
    \end{tabular}
    \caption{The model performance scores per NER tag. The evaluation is based on the separated test set.}
    \label{tab:tags_scores}
\end{table}

\begin{figure}[!ht]
    \centering
    \fbox{\includegraphics[width=0.98\textwidth]{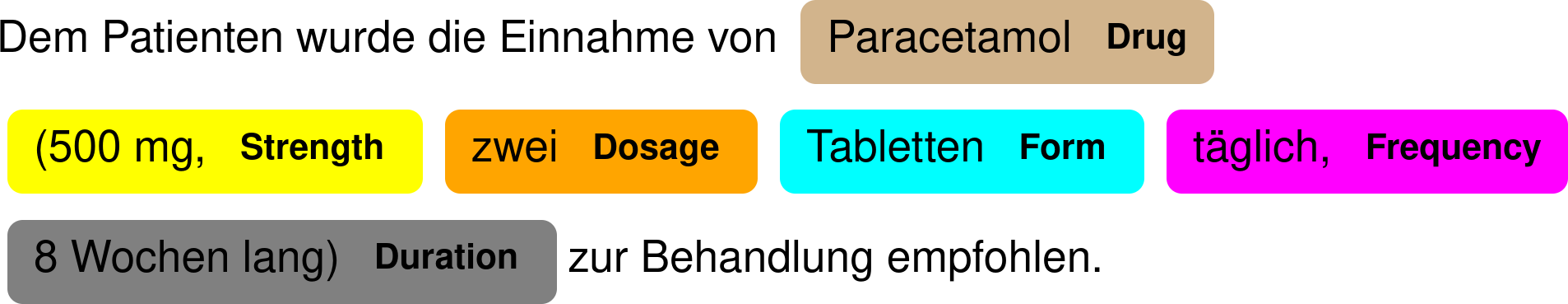}}
    \caption{Demonstration of successfully detected entities from German text}
    \label{fig:ner_demo}
\end{figure}

For demonstration purposes, a generic German sentence is shown in figure \ref{fig:ner_demo}. The annotations were inferred from the final model.

\section{Discussion}

In general, the availability of German NER models and methods for medical and clinical domains leaves much to be desired as described in previous sections. Analogous to that fact, German datasets in such domain are largely kept unpublished and are not available to the research community. However, its implications are significantly broader. In the case of unpublished NLP models, it renders independent reproduction of results and fair comparisons impossible. In the case of lacking datasets, novel competitive data-driven techniques cannot be developed or validated easily.

As a consequence, we cannot use such independent datasets for an extended evaluation of our model in order to estimate the inherent dataset bias of our custom dataset.

Regarding the topic of our custom dataset synthesis, one should emphasize that the outcome quality of the custom dataset and thus, the quality of the model, is likely to be influenced by the quality of the English-to-German translation engine. While in the case of English-to-German, modern NMT models are often sufficient and output reasonable results in the majority of text samples, the results are likely to worsen in the context of low-resource languages where powerful NMT models are not available.

The choice of the statistical model and the slim neural model architecture in particular is attributed to its small computational footprint while being able to achieve satisfying results. In addition, the NER pipeline of SpaCy explicitly induces inductive bias through hand-crafted feature extraction during the token embedding stage. However, since the focus of our work lies on the integration of NMT-based data for training purposes, we consider an exhaustive hyperparameter optimization as well as the utilization of a transformer-based model for improved NER performance scores as future work.

As this paper primarily focuses on the feasibility of training an NER model on synthesized datasets from public English datasets, we regard a deeper analysis of the obtained dataset corpus as future work.

\section{Conclusion}
In this paper, we presented the first neural NER model for German medical text as an open, publicly available model that is trained on a custom German dataset from an publicly available English dataset. We described the method to extract and postprocess texts from the masked English texts, and generate German texts by translating and cross-lingual token aligning. In addition, the NER model architecture was described and the final model performance was evaluated for single NER tags as well as its performance in total.

We believe that our model is a well-suited baseline for future work in the context of German medical entity recognition and natural language processing. The need for independent datasets in order to further improve the situation for the research community on this matter has been highlighted. We are looking forward to compare our model to upcoming German medical NER models.

The model as well as the training/test data are available at the following repository on GitHub: \url{https://github.com/frankkramer-lab/GERNERMED}.

\section*{Acknowledgment}
This work is a part of the DIFUTURE project funded by the German Ministry of Education and Research (Bundesministerium für Bildung und Forschung, BMBF) grant FKZ01ZZ1804E.

\bibliographystyle{plain}

\end{document}